\documentclass{article}

\usepackage{arxiv}

\usepackage[utf8]{inputenc} 
\usepackage[T1]{fontenc}    
\usepackage{hyperref}       
\usepackage{url}            
\usepackage{booktabs}       
\usepackage{amsfonts}       
\usepackage{nicefrac}       
\usepackage{microtype}      
\usepackage{lipsum}		
\usepackage{graphicx}
\usepackage{natbib}
\usepackage{doi}
\usepackage{amsmath}

\title{DEPTH ESTIMATION MAPS OF LIDAR AND STEREO IMAGES}

\author{
Fei Wu$^\dag$\\
School of Computer Engineering\\
The Australian National University\\
Canberra 2601 \\
\texttt{wufei.mlcv@gmail.com} \\
\and
Luoyu Chen$^\dag$\\
School of Computer Engineering\\
The Australian National University\\
Canberra 2601 \\
\texttt{luoyu.chen.mlcv@gmail.com} \\
}

\date{}

\renewcommand{\headeright}{Technical Report}
\renewcommand{\undertitle}{Technical Report}

\begin{document}
\maketitle
 \begin{figure}[h]
    \centering
    \includegraphics[width=1\linewidth]{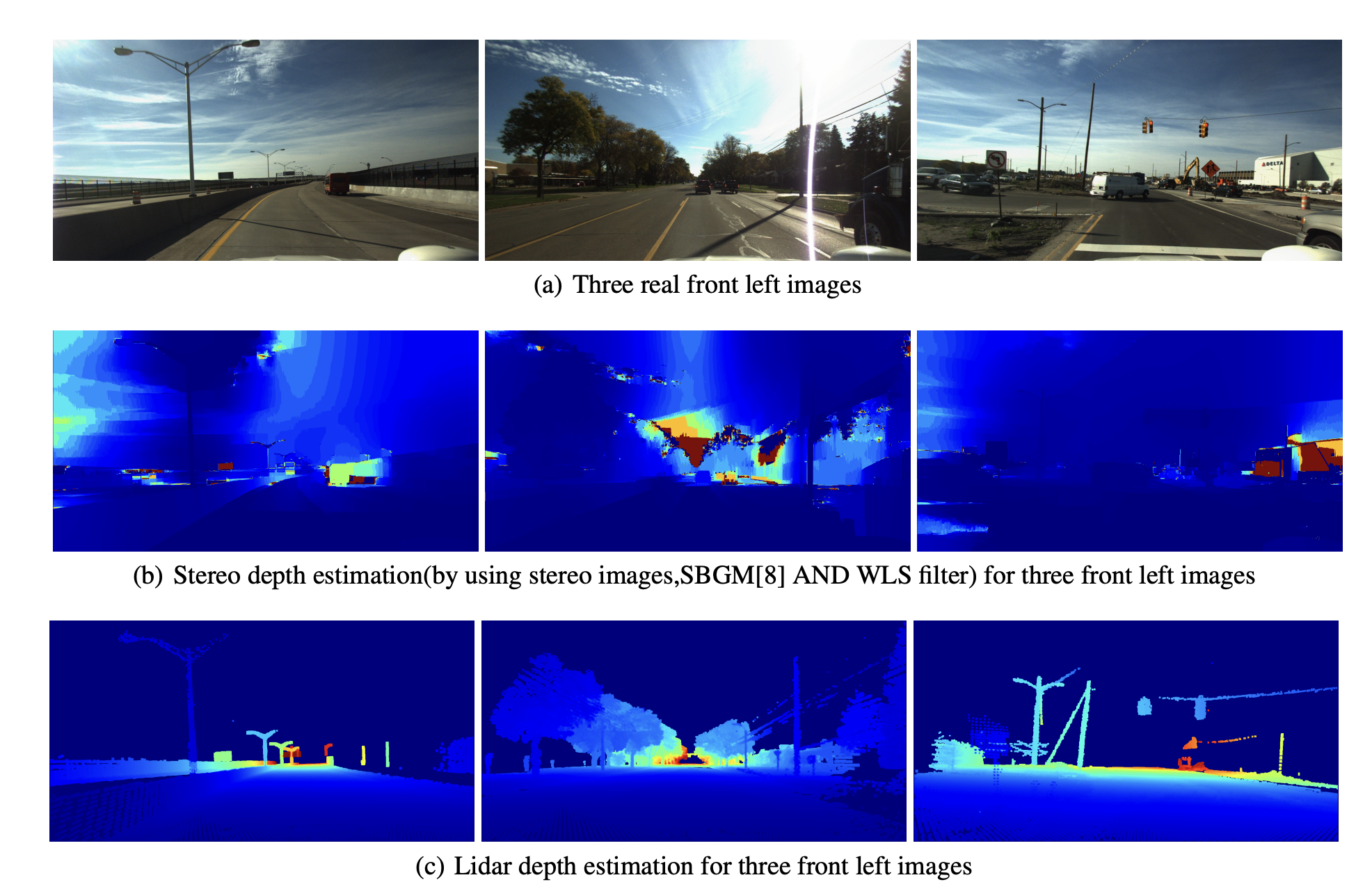}
    \caption{ First row (a) are showing three front left(fl) images(named fl1,fl2,fl3 images); Second row (b) are showing
stereo depth estimation for three front left images based on stereo images(front left and front right images); Third row(c)
is Lidar depth estimation (called lidar depth estimation in this report) by lidar data for three front left images}
    
    \label{fig:mesh2}
\end{figure}
\begin{abstract}
This paper is focusing on evaluation and performance about depth estimations based on lidar data and stereo images(front left and front right). The lidar 3d cloud data and stereo images are provided by ford. In addition, this paper also will explain some details about optimization for depth estimation performance. And some reasons why not use machine learning to do depth estimation, replaced by pure mathmatics to do stereo depth estimation. The structure of this paper is made of by following:\textbf{(1) Performance:} to discuss and evaluate about depth maps created from stereo images and 3D cloud points, and relationships analysis for alignment and errors;\textbf{(2) Depth estimation by stereo images: }to explain the methods about how to use stereo images to estimate depth;\textbf{(3)Depth estimation by lidar: }to explain the methods about how to use 3d cloud datas to estimate depth;In summary, this report is mainly to show the performance of depth maps and their approaches, analysis for them.
\end{abstract}
\def\thefootnote{$\dag$}\footnotetext{These authors contributed equally to this work}

\keywords{Depth estimation \and Lidar \and Stereo images\and Machine learning}
 \begin{figure}[h!]
    \centering
    \includegraphics[width=1\linewidth]{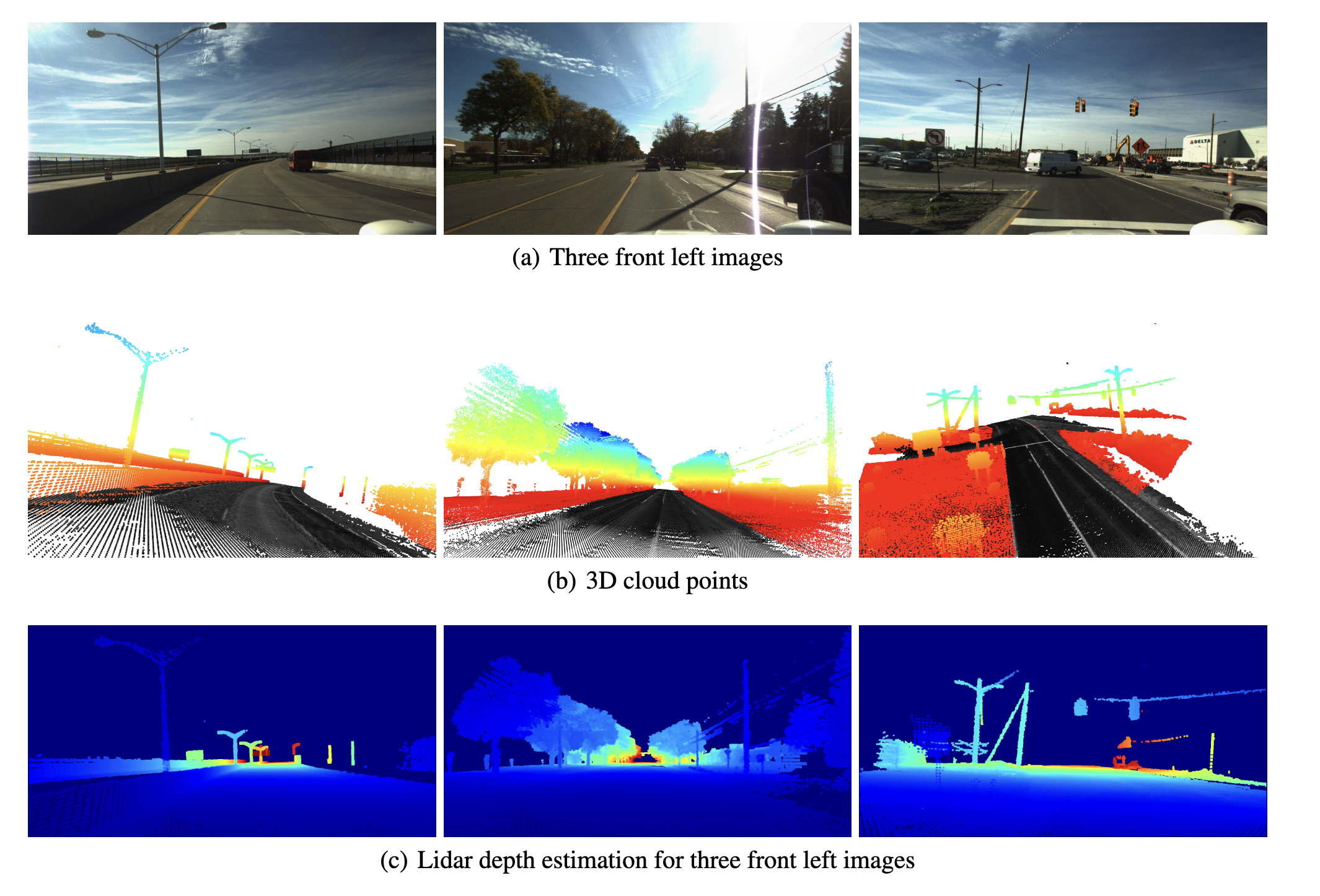}
    \caption{ First row (a) are showing three front left(fl) images(named fl1,fl2,fl3 images); Second row(b) is 3d clouds data,  Third row(c) is Lidar depth estimation respectively for three fl1 fl2 fl3 front left images}
    \label{fig:mesh2}
\end{figure}
\section{Performance}
The Fig1 is showing the Lidar depth estimation from ford Lidar data and stereo depth estimation by stereo images(front left and right images). Fig1.1st row are front left images, Fig1.2nd row are stereo depth estimation for front left images based on stereo images(front left and right images) doing by pure mathematics and its disparity maps. The Fig1.3ird row is lidar depth estimations for these front images from lidar data(3D cloud points).\\ 
\newline
From Fig1, comparing performances of  stereo depth estimation and Lidar depth estimation, Lidar depth estimation performance is showing better and has more accuracy. Because 3D points data quality is so high and can't be affected and influenced by outdoor environment. However, using 3D cloud points data provided by ford to do Lidar depth estimation doesn't have information about dynamic objects(i.e cars , pedestrian,sky and etc) but can provide information about static objects. For stereo depth estimation by pure mathematics approach, stereo images quality is so important and is easily affected by outdoor environment( for instance  clouds, shadows and high lightness, in Fig1.1st row, the mid image has a high lightness which affect the image quality.). But using stereo images can provide the more information about dynamic objects(i.e cars , pedestrian,sky and etc). Lidar depth estimation with better performance than stereo depth estimation can be used as ground truth.  

\subsection{Lidar depth estimation and 3d cloud points alignment}
Based on Fig2, Fig2(b) is 3D cloud data in global frame showing around environment(i.e road, trees, polar and etc), compared with Fig2(c) Lidar depth, they have good alignment(polars, brands, shapes of trees) which ensure some key characteristics from front left images and some statistic objects scanned by lidar, all of them are correctly projected to our front left image as for Lidar depth estimation. And for Fig2(c) lidar depth maps clearly show the positions of relationships among different objects.

\subsection{Lidar depth estimation without depth completion}
The Fig.3 is about comparison of lidar depth with completion and lidar depth without completion. The lidar depth with completion is same with Fig1(c) and Fig2(c), all of them are showing same performance. 3D cloud points from Lidar prjecting to a 2D image will face a problem about sparse points in short distance, dense points in long distance. In Fig3(a) is performance without depth completion, Fig3(b) is with depth completion. We can see some details of object will recovery and fill some black holes, after using depth completion. However, importantly, besides of doing depth completion optimization, performance of Fig3(b) also did the other optimization called saving characteristic optimization(see 1.3 section Lidar depth estimation without saving key characteristics).
 \begin{figure}[h!]
    \centering
    \includegraphics[width=1\linewidth]{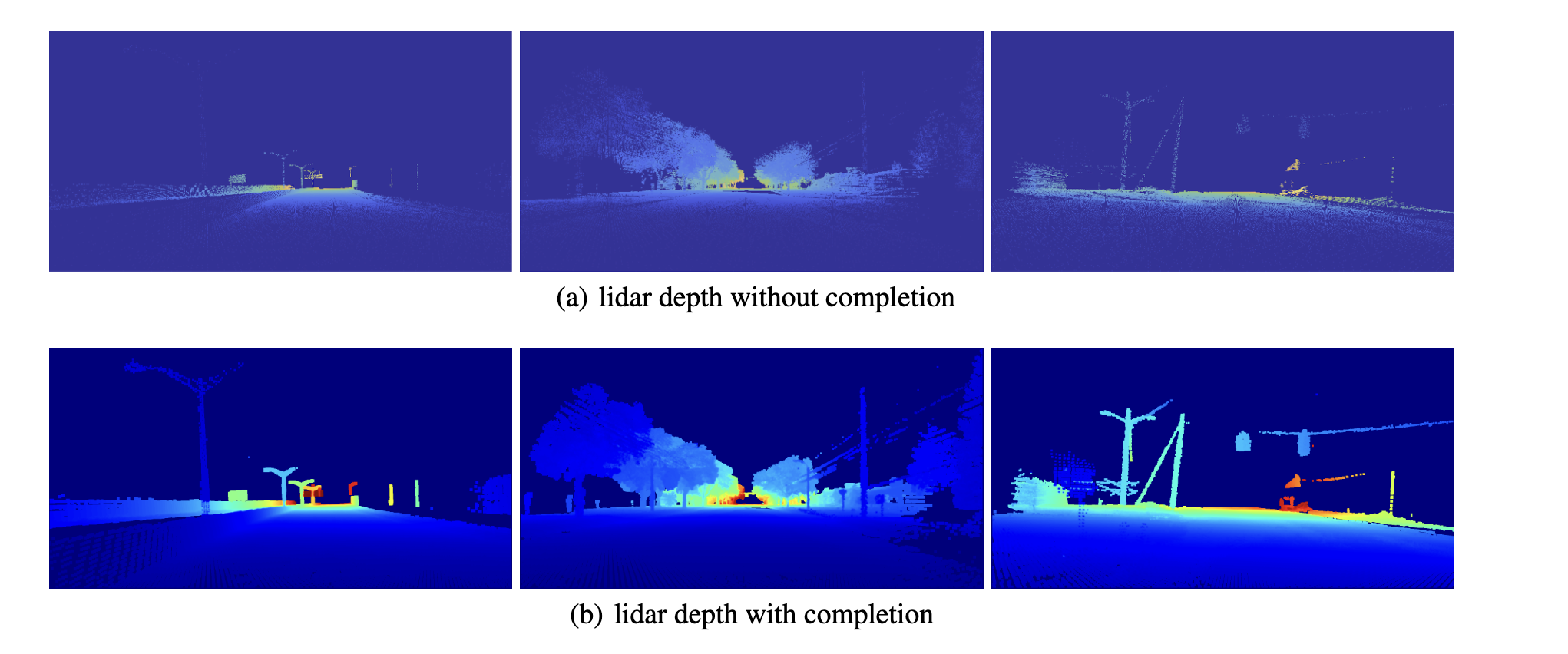}
    \caption{ First row (a) are showing three front left(fl) images(named fl1,fl2,fl3 images); Second row(b) is 3d clouds data,  Third row(c) is Lidar depth estimation respectively for three fl1 fl2 fl3 front left images}
    \label{fig:mesh2}
\end{figure}
\subsection{Lidar depth estimation without saving key characteristics}
As we can see if we don’t save characteristic, the bad performance will be like in Fig.4(a)(c), after using depth completion, the background scene will cover front scene, even if before doing depth completion and you optimize it by low depth covering high depth in same pixel, you also will get such bad performance. The reason is that depth completion by pure math approach will do more completion for dense depth pixel than for sparse depth pixel. So my approach is to enhance characteristic before doing depth completion. Before doing depth completion, spread sparse depth pixel(short distance to camera) to around 3x3 neighbors.  
 \begin{figure}[h!]
    \centering
    \includegraphics[width=1\linewidth]{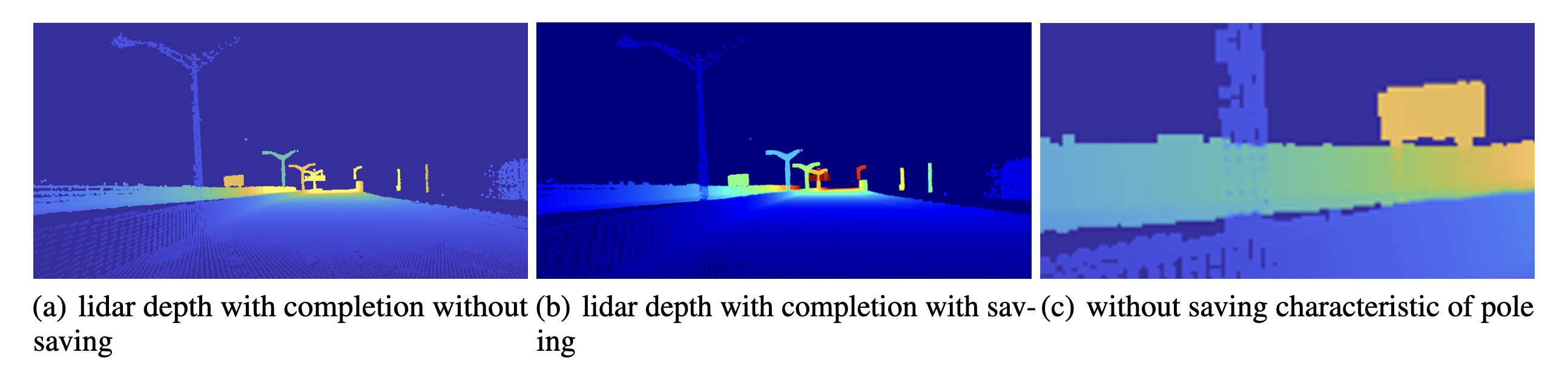}
    \caption{ (a) is using depth completion but no using saving key characteristics, (b)is using depth completion and using saving key characteristics which is my best performance of lidar depth estimation. (c) is to show that if we not save key characteristics, depth completion also will make its background pixel(fence) cover front pixel(pole)}
    \label{fig:mesh2}
\end{figure}
\subsection{Lidar depth and real images alignment}
We can see the Fig.5 is about Lidar depth and real images alignment. We can see the position almost same and they have good alignment.From Fig5.(c), we can see the positions for depth of each pixel are almost in correct positions to real image, but it also has little bias. It shows our lidar depth map about positions are almost correct in positions. In Fig5(c), the light blue dots are lidar depth pixels. 
 \begin{figure}[h!]
    \centering
    \includegraphics[width=1\linewidth]{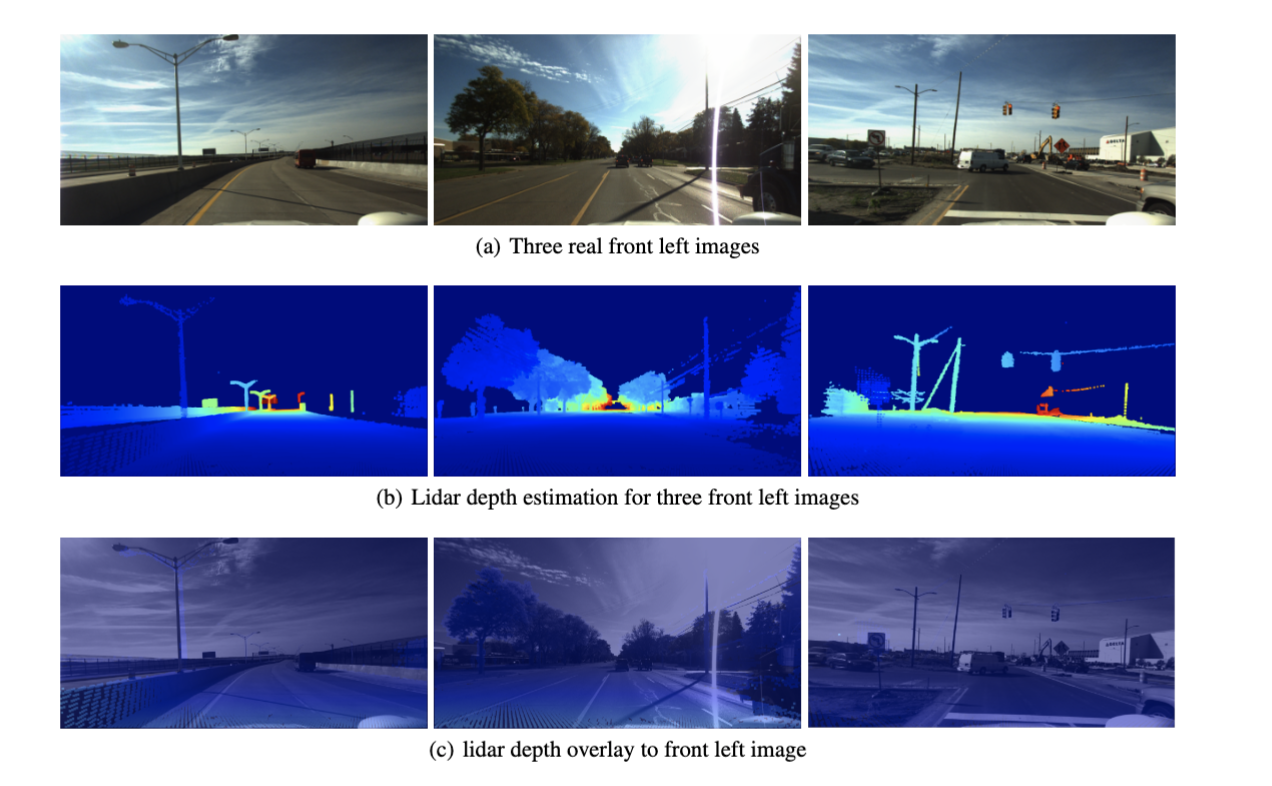}
    \caption{ First row (a) are showing same images with Fig1(a); Second row is lidar depth map; Third row is lidar depth map overlay to real image}
    \label{fig:mesh2}
\end{figure}
\subsection{Stereo depth from derivative maps based on stereo images}
The stereo depth is based on disparitive maps from stereo images(left and right camera images), showing in Fig.6. It is based on formular $depth=\frac{len\times baseline}{disparitives}$, where len=945.391406, baseline=0.5764. It used the pure math, by matching block in a polar line and compare to get disparities(pixel), importantly, the depth maps are in int16. We can see, it is easily affected by image quality, for instance high sun light, shadows and complex scene(clouds,trees and cars).
 \begin{figure}[h!]
    \centering
    \includegraphics[width=1\linewidth]{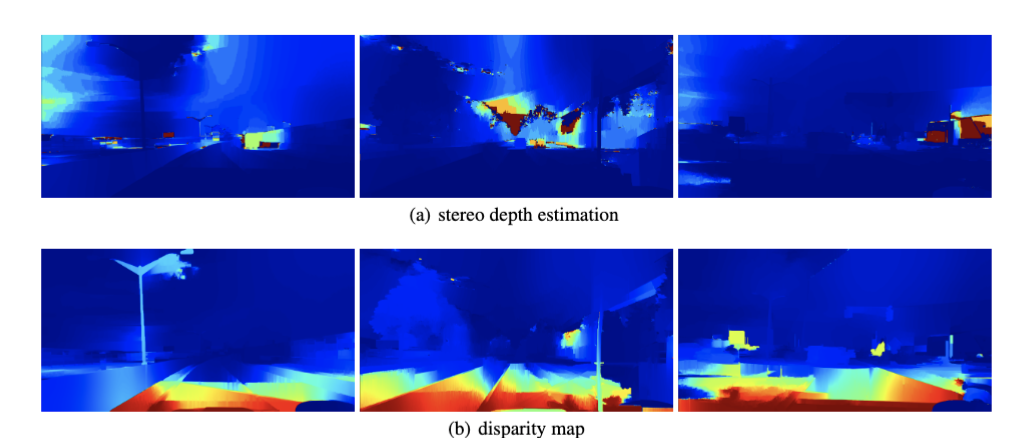}
    \caption{ (a) is stereo depth((by using stereo images,SBGM[8] AND WLS filter)) and it is same images with Fig1(b);(b) are respectively disparities maps}
\end{figure}
\subsection{Lidar depth and stereo depth alignment}
The Fig.7 is showing Lidar depth and stereo depth alignment.The background is stereo depth, the red points are lidar depth. As we can see the positions of objects are all correct. By the way our image shape is 860x1656\\
 \begin{figure}[h!]
    \centering
    \includegraphics[width=1\linewidth]{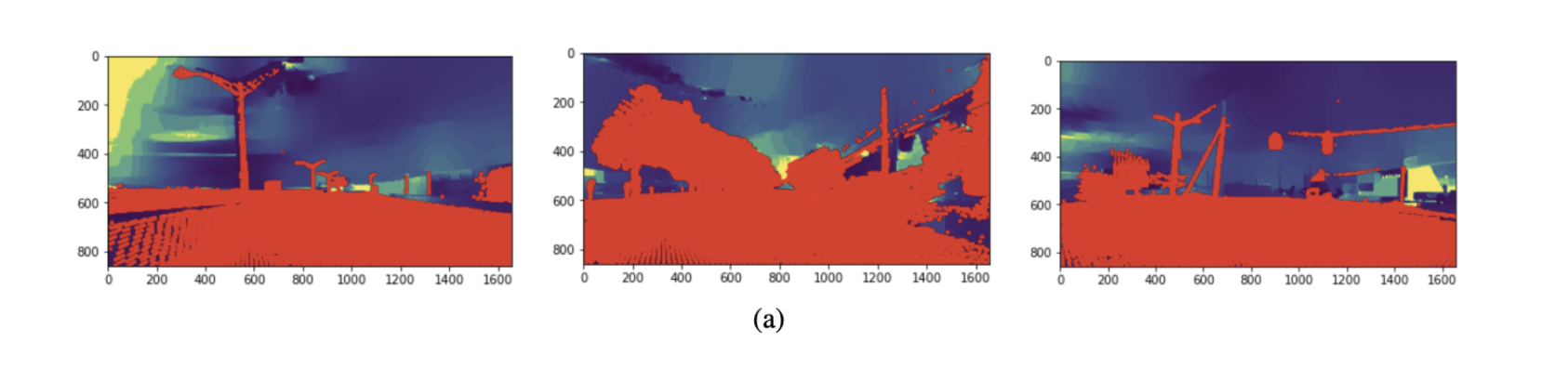}
    \caption{ Lidar depth maps overlay to stereo depth maps, red dots are from lidar depth maps, background images are
stereo images}
    \label{fig:mesh2}
\end{figure}
\subsection{Errors of stereo depth and lidar depth}
Compare error of stereo depth estimation Fig1(b) and lidar depth Fig1(c) as gt, for fl1,fl2, and fl3 front left images, their errors are 5.448741012245815, 12.107053982698574 and 4.933980030333671 average error, which means for each pixel, they have different avg meter errors, we can see for fl2 real image, they have more avg meter error for each pixel, the reason is fl2 image have more complex scene(high sun lightness and trees). 
\subsection{Performance of pretrained models for disparities}
In here, we try to use some models(neural network), trained by kitti to get . They are: LEAstereo, Lac-GwcNet, Onnx-Hinet stereo,PyTorch-High-Res-Stereo-Depth-Estimation. The Fig.8 is the performance disparitives maps of them about Kitti. The Fig.9 is using these models to folder dataset for getting their respectly derivatives performance, as we can see, higher rank in kitti, worse performance in folder, which mean these models almost are overfitting\\
 \begin{figure}[h!]
    \centering
    \includegraphics[width=1\linewidth]{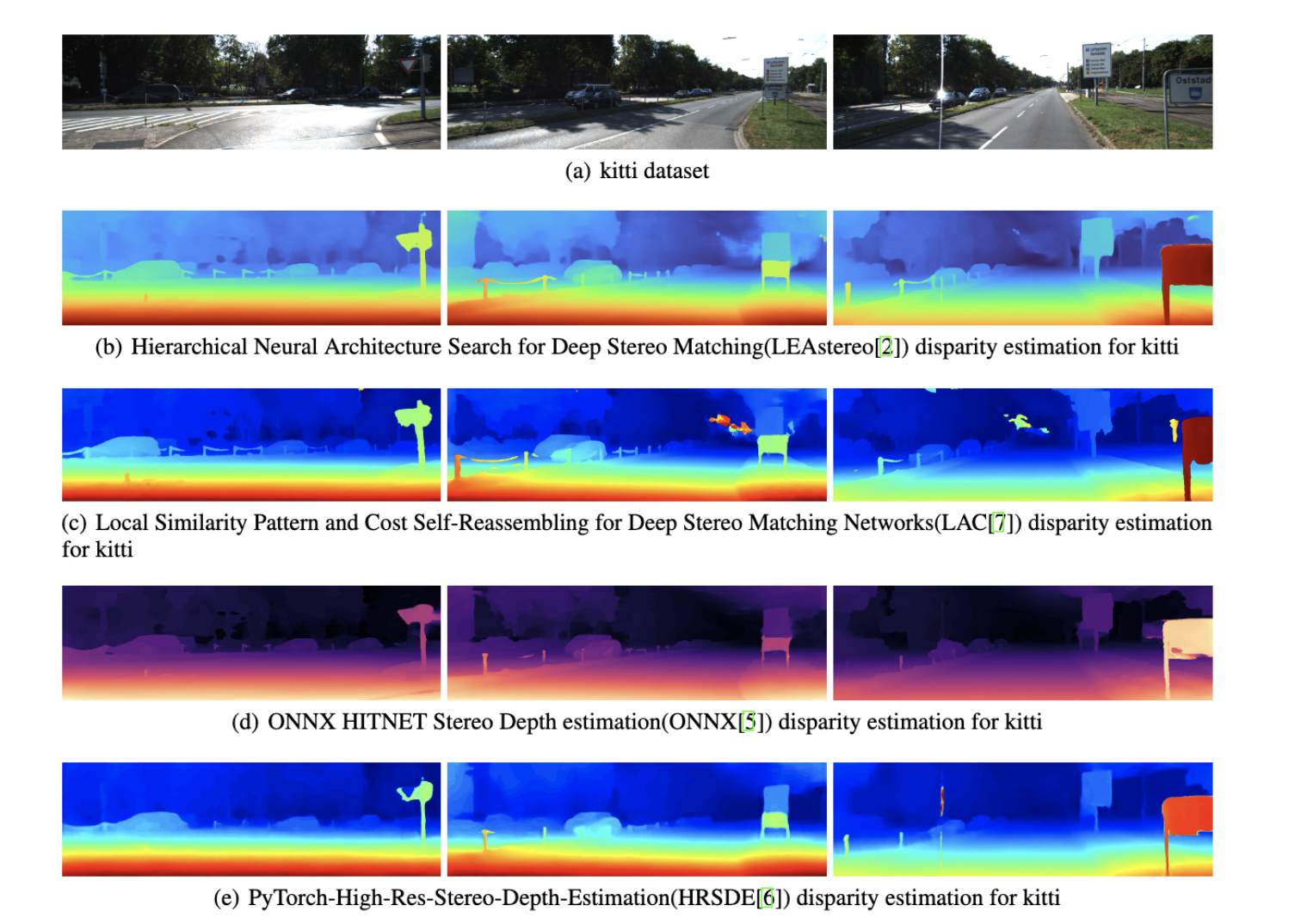}
    \caption{ a) are real kitti dataset; (b) is LEAstereo[2] method for disparity estimation; (c) is LAC[7] method for
disparity estimation ; Fourth row is ONNX[5] method for disparity estimation; Fifth row is HRSDE[6] method for
disparity estimation}
    \label{fig:mesh2}
\end{figure}
\subsection{Depth maps from HRSDE disparity}
Fig.10 is about the depth maps from disparity maps of kitti trained hrsd model.The errors (compared with lidar depth) for each pixel about depth maps from  hrsde disparity map to three images respectively are, 16.334722626939055, 24.078121162484532 and 13.90534372304212, which is larger than the errors of lidar depth and stereo depth(math). In summary, i think using mathmatics to get depth stereo from disparity performance is better than using kitti trained model to do this\\
 \begin{figure}[h!]
    \centering
    \includegraphics[width=1\linewidth]{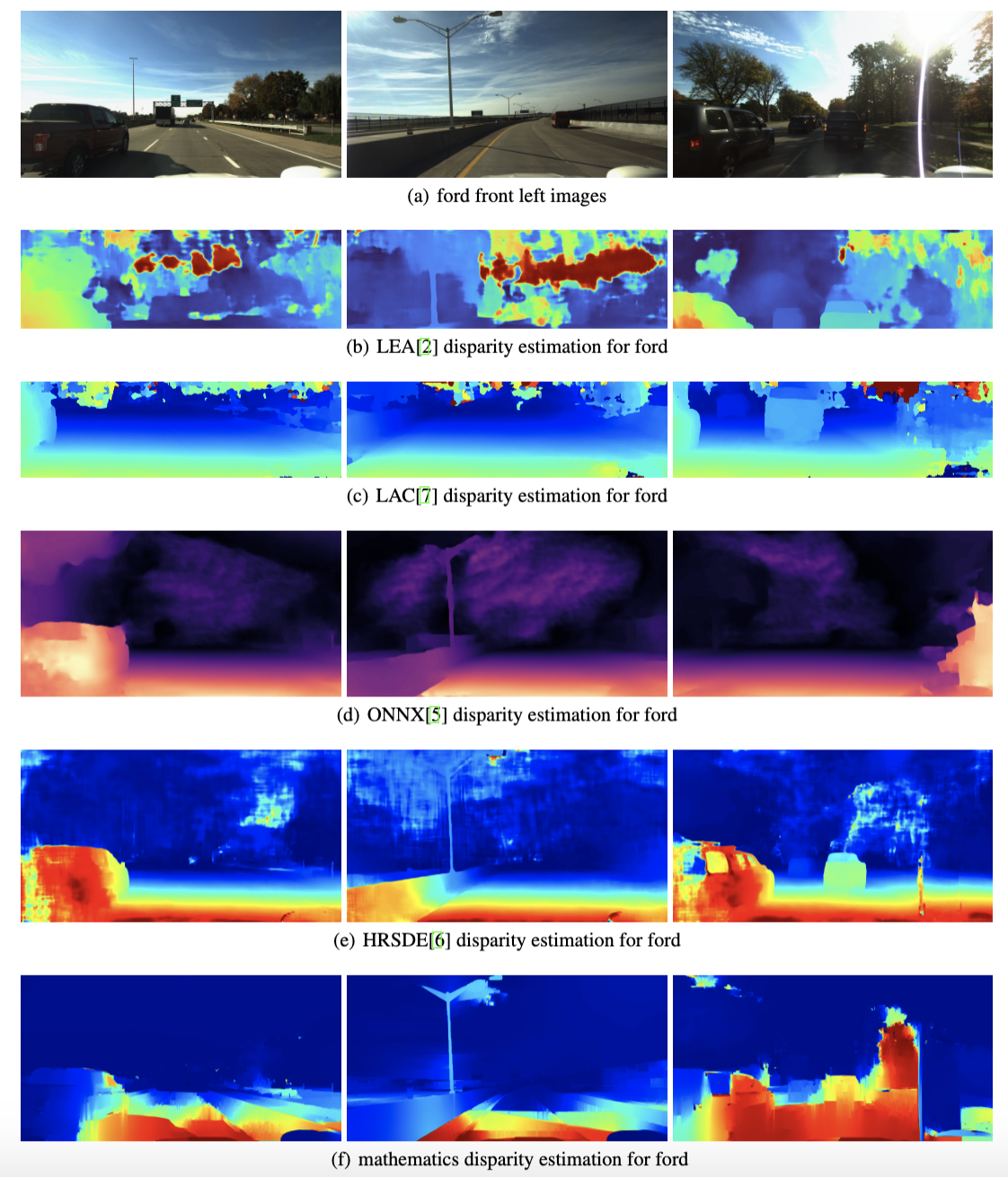}
    \caption{ (a)Disparities performance for ford dataset; (b) is trained kitti for LEA[2] method used for ford; (c) is trained
kitti for LAC[7] method used for ford; (d) is trained kitti for ONNX[5] method used for ford; (e) is trained kitti for
HRSDE[6] method used for ford; (f) row is disparity of stereo images by mathmatics which is same images of Fig6(b)}
    \label{fig:mesh2}
\end{figure}

\begin{figure}[h!]
    \centering
    \includegraphics[width=1\linewidth]{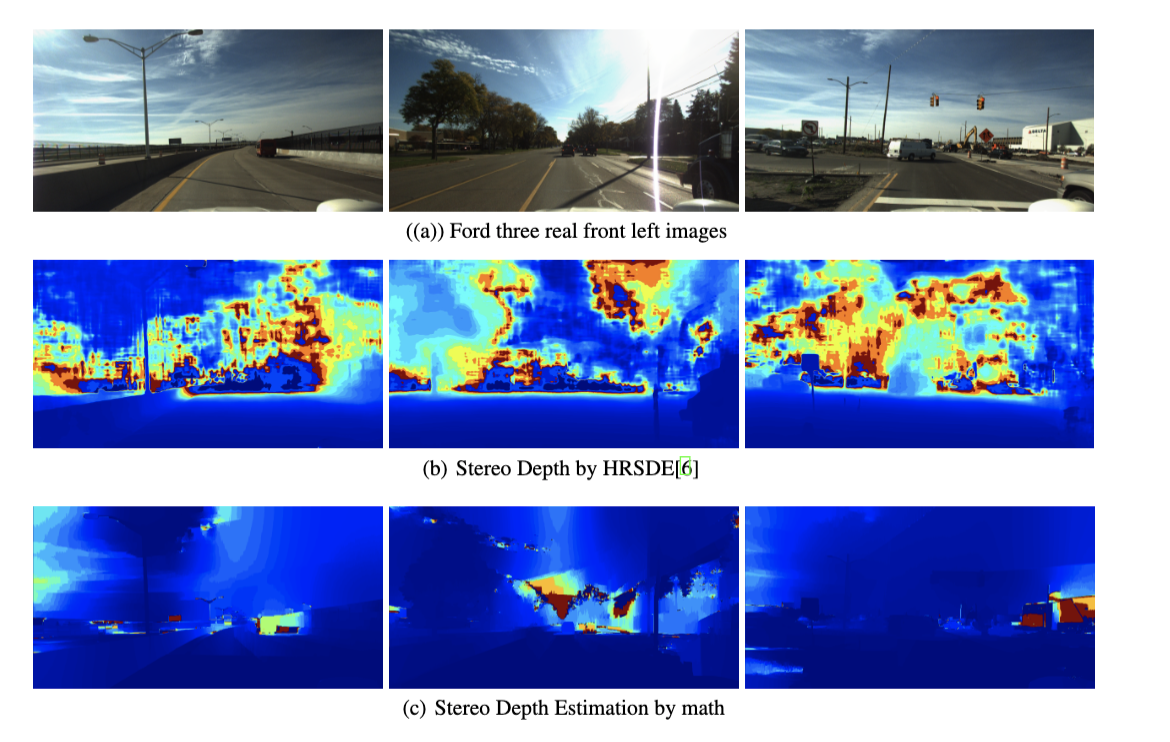}
    \caption{ a) are front left images; (b) is depth maps from HRSDE[6] disparity map; (c) is stereo depth from disparity
maps based on mathmatics which are same images with Fig1(b)}
    \label{fig:mesh2}
\end{figure}

\section{Depth estimation by stereo images}
Basic idea is on Fig.11 about binocular stereo, in a polar line(scan line) to match almost same pixel in both stereo images(left and right), then get disparity position $d=u-u'$, then get depth $Z=\frac{len\times baseline}{d}$.In here, we will mainly need to use cv2 package for codes on python. Firstly, we should use gaussianblur to smooth stereo images. Then we should create a matcher (cv2.StereoSGBM\_create) which is to create block to match pixels. Acturally, there are so many algorithms about matching, importantly they are BM(block matching) and SGBM(Semi Global Block Matching), their basic ideas are using block to scan on polar lines to calculate SAD(sum absolute difference),then, it calculates the similarity
between different windows(blocks) and this window in the traversal, and takes the most similar window as the
final result. BM is simple just cv2.StereoBM\_create(numDisparities,block\_size), where numDisparities is length of disparities(16times) to scan or match in such length, maxDisparity=minDisparity+numDisparities-1 \\

SGBM is local matching algorithm which is fast and simple, but the disparity map obtained by experiments is
rough and the algorithm is not robust. It is easily affected by external factors like illumination and
occlusion. The global matching algorithm has high precision, so its complexity is also high. SGBM
algorithm is a semi global stereo matching algorithm. By selecting the difference of each pixel,
SGBM forms a disparity map and sets a global energy function related to it, so as to minimize the
energy function below.
$$E(d)=\sum_pC(p,d_p)+\sum_{q\in N_p}P_1T[|d_p-d_q|]+\sum_{q\in N_p}P_2T[|d_p-d_q|>1]$$ 
Where $\sum_pC(p,d_p)$ is sum of all pixel matching costs based on disparity d. It has two energy parameters($P_1,P_2$) to punish disparity values for more accuracy, others almost same with BM operations. In summary SGBM performance is better than BM, but SGBM has worse computation than BM.The Fig.12 is the performance  disparity of BM and SGBM.
We can see the performance is not good, even you smoothed stereo images and normalize them. Then, we should introduce a weighted least squares filter(WLS filter) to save the edges (enhance high gradient) and smooth some place with low gradient. Set original image as $g$, filtered image as $u$, then the loss function $f(u)$:
$$\sum_p((u_p-g_p)+\lambda(a_{x,p}(g)(\frac{\partial u}{\partial x})^2_p)+a_{y,p}(g)(\frac{\partial u}{\partial y})^2_p)$$
minimize such loss function to get filtered image $u$ from original image $g$. But in codes, cv2 provided an easy way, firstly to get wls filter based on sgbm matcher size and settings (wls\_filter). Then based on above loss function, we can get $u=(I+\lambda a L)^{-1}g$, $g$ is original image. $\lambda$ is to balance between the data term and the smoothness term. Increasing $\lambda$ will produce smoother images. $a$ as $\alpha$ is control over the affinities by non-lineary scaling the gradients. Increasing alpha will result in sharper preserved edges. $L$ is source image for the affinity matrix,Same dimensions as the input image, which acturally is Laplace homogenous matrix. In final we will get good performance disparity,Fig.1, second row. In final, the code is filtered\_disp = wls\_filter.filter(left\_disp, left\_image(original image, in here it is smoothed image), disparity\_map\_right=right\_disp);\\

There is a optimization for its final performance about Fig13 which has no matching columns problem. Adding or padding an extra column(hight $\times$ numDisparity) can solve this problem(my optimization). The reason why have such problem is that during left to match right, there is no pixel matching.
\begin{figure}[h!]
    \centering
    \includegraphics[width=1\linewidth]{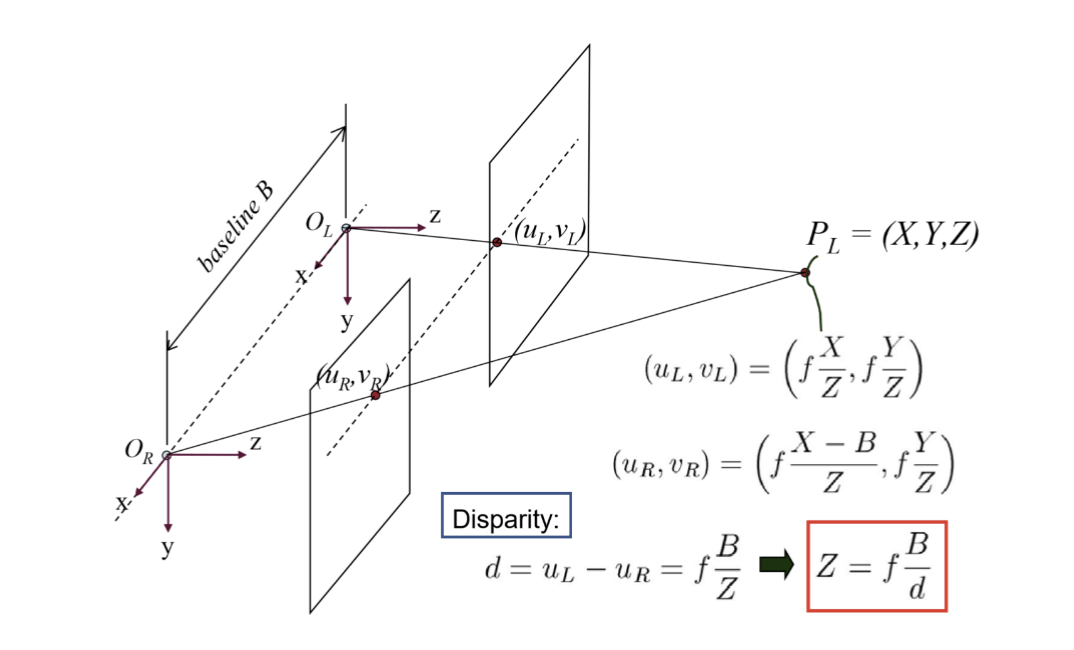}
    \caption{ cisparity and depth}
    \label{fig:mesh2}
\end{figure}
\begin{figure}[h!]
    \centering
    \includegraphics[width=1\linewidth]{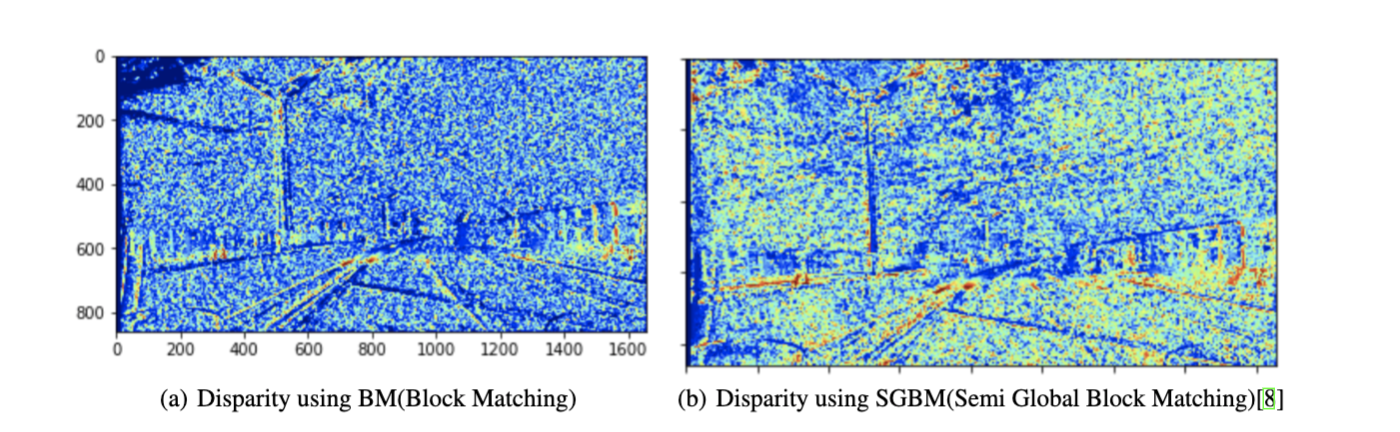}
    \caption{ Disparity performance, left is using BM, right is using SGBM[8]}
    \label{fig:mesh2}
\end{figure}
\section{Depth estimation by lidar}
(1)Find your target stereo images, and use their name(frame time) to get their GT pose(pose position x,y,z). Then use pose positions to find 3d cloud points and grav points(world frame)
(2) Change world frame to body frame, based on GT pose(pose position and quaternion), importantly, quaternion can be changed to rotation matrix:
$$^{G}R_B=\begin{bmatrix}1-2y^2-2z^2&2xy-2zw&2xz+2yw\\
2xy+2zw&1-2x^2-2z^2&2yz-2xw\\
2xz-2yw&2yz+2xw&1-2x^2-2y^2\\
\end{bmatrix}$$
transposition rotation matrix and minus transposition rotation matrix multiply trans. In summary, the function about changing points of global frame to points of body frame is
$$^{B}_{B}P_{G}=^{B}R_{G}{^{G}_{G}P_{G}}-^{B}R_{G}{^{G}_{G}t_{B}}$$
where $P$ means the 3D point, $B$ means under body frame, $G$ means under global frame, and $^{B}R_{G}=(^{G}R_{B})^T$, note: generally observer frame is on original frame\\
(3) same operations to change body frame to camera frame, following function:
$$^{C}_{C}P_{B}=^{C}R_{B}{^{B}_{B}P_{B}}-^{C}R_{B}{^{B}_{B}t_{C}}$$
where $C$ means under camera frame\\
(4) Using projection matrix to project into 2D image, depth value will be pixel value.
\subsection{Optimization about saving characteristics}
To solve the problem about Fig4, we also should increase some low depth pixel to spread around.In here, I spread them to 4x4 range.\\
\begin{figure}[h!]
    \centering
    \includegraphics[width=1\linewidth]{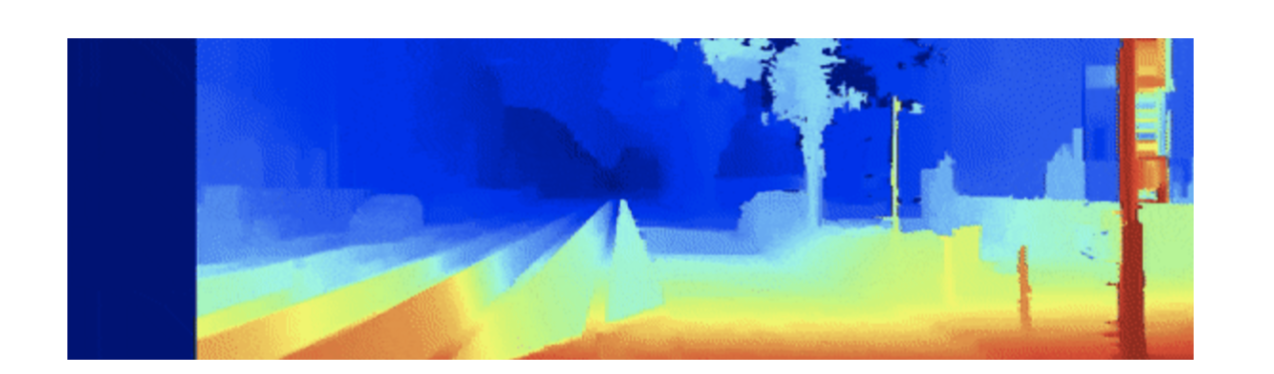}
    \caption{ No matching columns(left deep blue bar, almost equal to numDisparity which is your scan mataching range),
after my optimization, this problem also is addressed, then we get final performance stereo depth estimation Fig1(b)}
    \label{fig:mesh2}
\end{figure}
\subsection{Depth completion}
Because to project in to 2D, there are lots of round(), some pixel will lost, and lidar 3d points are too sparse in 2d image.Easily to say, for instance, if ngrid=4(neighbors), it means we used 9x9 neighborhood and we calculated weighted depth information according to the distance of the neighborhood. Basically, we used the distance of the neighborhood as a weight of sum process.For instance, some pixel is 0 depth which means no information from lidar 3d and will be sparse, then we should see their neighbors and sum the weight(depth/distance), then in final become depth from weight multiply sum distance from their neighbors~\cite{lv2020experiment}. The whole project is made of matlab and python, remember to use h5 file for saving data, importantly, h5 from python to matlab should transposition matrix data.
\begin{figure}[h!]
    \centering
    \includegraphics[width=1\linewidth]{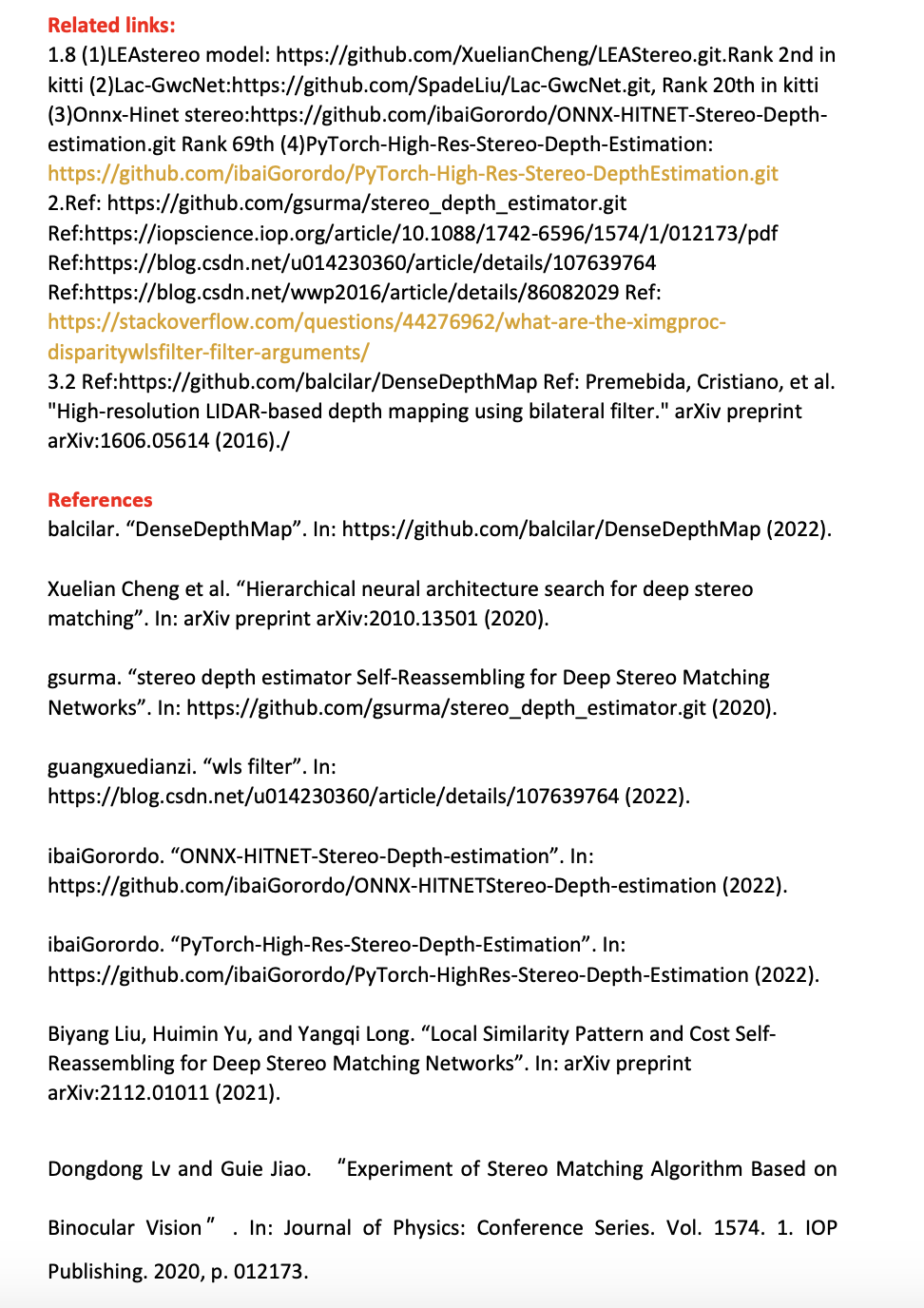}

\end{figure}

\end{document}